\pdfoutput=1

\documentclass[11pt]{article}

\usepackage[preprint]{acl}

\usepackage{times}
\usepackage{latexsym}
\usepackage{amsmath}
\usepackage{amssymb}
\usepackage[T1]{fontenc}

\usepackage[utf8]{inputenc}

\usepackage{microtype}

\usepackage{inconsolata}

\usepackage{graphicx}

\usepackage{makecell}
\usepackage{multirow}

\usepackage{booktabs}

\usepackage[exponent-product=\cdot]{siunitx}

\usepackage{cleveref}

\title{BabyLlama-2: Ensemble-Distilled Models Consistently Outperform Teachers With Limited Data}

\author{Jean-Loup Tastet \\
  University of Copenhagen \\
  Department of Computer Science \\
  Copenhagen, Denmark \\
  \texttt{jeta@di.ku.dk} \\\And
  Inar Timiryasov \\
  University of Copenhagen\\ 
  Niels Bohr Institute\\
  Copenhagen, Denmark\\
  \texttt{inar.timiryasov@nbi.ku.dk} \\}

\begin{document}
\maketitle
\begin{abstract}
We present BabyLlama-2, a 345 million parameter model distillation-pretrained from two teachers on a 10 million word corpus for the BabyLM competition. On BLiMP and SuperGLUE benchmarks, BabyLlama-2 outperforms baselines trained on both 10 and 100 million word datasets with the same data mix, as well as its teacher models. Through an extensive hyperparameter sweep, we demonstrate that the advantages of distillation cannot be attributed to suboptimal hyperparameter selection of the teachers. Our findings underscore the need for further investigation into distillation techniques, particularly in data-limited settings.
\end{abstract}

\section{Introduction}

With frontier model training runs using beyond $10^{25}$ FLOPs~\citep{dubey2024llama}, training efficiency has become a billion-dollar question. 
Humans are vastly more sample efficient than current Large Language Models (LLMs). For example, a typical 13-year-old child has been exposed to less than $100$ million words (extrapolating from \citet{mapping-language-development}), whereas Llama-3.1 has been trained on $15.6$ trillion text tokens. The goal of the BabyLM Challenge~\citep{babylm-2024} is to optimize pretraining given dataset limitations inspired by human development.

In this work, we present our contribution to the BabyLM challenge (Strict-Small Track), with the following main results:
\begin{itemize}
\item BabyLlama-2 model: This 345M decoder-only model\footnote{It is worth noting that encoder models are better suited for the evaluation tasks of the challenge than decoder ones. In last year's evaluation~\citep{warstadt2023papers}, the 125M parameter RoBERTa-base~\citep{liu2019roberta} performed on par with the 70B parameter Llama-2~\citep{touvron2023llama}. However, our focus throughout this paper shall be on generative, decoder models.}, distillation-pretrained on 9.5M words, outperforms baseline models trained on both 10M and 100M words (using the same data mix). It also surpasses similar models pretrained using conventional methods.

\item Extensive hyperparameter sweep: We have conducted a comprehensive hyperparameter optimization and demonstrated that distillation pretraining consistently outperforms the best models from the sweep.

\item Correlation between test loss and performance: As a byproduct of our sweep, we have identified a correlation between zero-shot performance on the BLiMP task and the model's test loss.
\end{itemize}
The success of distillation pretraining, i.e.\ pretraining from scratch with distillation loss, in our experiments highlights its potential as a powerful technique for improving model performance, especially in data-limited settings. While our findings are promising, they also raise intriguing questions about the nature of knowledge distillation and its interaction with pretraining objectives. Further investigation into these areas could yield valuable insights for the development of more sample-efficient language models.

\section{Related Work}
The first edition of the BabyLM challenge, which aims to optimize language model pretraining under data constraints inspired by human language acquisition, prompted numerous works on sample-efficient pretraining. For a detailed summary of all contributions, see the review by \citet{warstadt2023papers}. 

Outside the BabyLM context, relatively few works address training on limited language datasets. Notable exceptions include 
\citet{2023arXiv230516264M}, who studied the scaling of data-constrained LLMs. Their main finding is that training for more than 4 epochs leads to diminishing returns.
\citet{luukkonen2023fingpt} trained FinGPT on more than 30B tokens in Finnish language. Although resource-constrained, this dataset is significantly larger than that of the BabyLM Challenge.
A sample-efficient modification of BERT architecture was proposed by~\citet{ltg-bert}, with a model trained on a 100M word dataset from the British National Corpus outperforming the original BERT model.

The existing literature on training small models often focuses on models deployable on edge devices, such as MobileLLM \citep{liu2024mobilellm}. However, these works typically concentrate on deployment efficiency rather than sample efficiency.

Knowledge distillation has recently attracted significant attention, primarily for deployment efficiency reasons (see \citet{xu2024survey} for a systematic review). Typically, this involves using large frontier models as teachers to train smaller student models. In contrast, BabyLlama-2 utilizes distillation for sample-efficient pretraining, using similar-sized teacher models trained on the same limited dataset.

\section{Background}
Knowledge distillation, introduced by~\citet{2015arXiv150302531H}, is a technique for transferring knowledge from a "teacher" model to a "student" model. The core idea is to train the student to mimic the logit distribution (soft targets) produced by the teacher, rather than just the hard labels of the training data.
The distillation loss combines the standard cross-entropy loss with the soft target loss:
\begin{multline}
\mathcal{L}_{\text{distill}}(y, z_s, z_t) =
\alpha \,\mathcal{L}_{\text{CE}}(y, \sigma(z_s)) + \\
(1-\alpha)\, T^2 \, D_\text{KL}\left( \sigma(z_t/T)\, ||\, \sigma(z_s/T)\right)
\label{loss}
\end{multline}
where $\alpha$ balances the usual cross-entropy loss $\mathcal{L}_{\text{CE}}$ and the soft targets loss, $T$ is the temperature parameter that softens the probability distributions, $z_s$ and $z_t$ are respectively the logits of the student and teacher models, $\sigma$ is the softmax function, and $ D_\text{KL}$ denotes the Kullback-Leibler divergence. In our implementation, we use the averaged logits of an ensemble of teacher models as $z_t$. Moreover, unlike typical applications, our student and teacher models are of the same size.

\section{Model}
\paragraph{Architecture.}
Previous experiments have shown that the Llama architecture~\citep{llama}, featuring RoPE and a SwiGLU non-linearity, requires fewer epochs to reach minimal loss compared to GPT-2 or GPT-J architectures~\cite{timiryasov2023speed}.
After training a family of Llama models ranging from 16M to 728M parameters, we converged on a specific 345M model architecture 
suggested in MobileLLM~\citep{liu2024mobilellm} and also used in
 SmolLM~\cite{smollm2023}, whose hyperparameters are listed in \cref{tab:model_architecture}.
This design incorporates Grouped-Query Attention (GQA) and prioritizes depth over width.
Some details of our model selection are listed in appendix~\ref{sec:family}.

\begin{table}[h]
\centering
\begin{tabular}{ll}
\toprule
\textbf{Hyperparameter} & \textbf{Value} \\
\midrule
Vocabulary size & 16,000 \\
Number of layers & 32 \\
Number of heads & 15 \\
Number of KV heads & 5 \\
Embedding dimension & 960 \\
Hidden dimension & 2560 \\
Total parameters & 345M \\
\bottomrule
\end{tabular}
\caption{BabyLlama-2 Model Architecture.}
\label{tab:model_architecture}
\end{table}
\paragraph{Pretraining Approach.}
The particularity of the BabyLlama-2 model is to be distilled from an ensemble of teacher models, using the distillation loss~\eqref{loss}.
The teacher models share the same architecture and are pretrained on the same dataset using the standard cross-entropy loss.
The student model is then pretrained with the same hyperparameters, using the mean teacher logits $\bar{z}_t$ in the distillation loss $\mathcal{L}_{\mathrm{distill}}(y, z_s, \bar{z}_t)$.
For the 345M model, we observe diminishing returns when using more than two teachers.

\section{Experimental Setup}
\label{sec:experimental-setup}

\paragraph{Dataset.} We use the 10 million word BabyLM-2 dataset \citep{babylm-dataset}, that we split into 9.5M train and 0.5M validation splits, as well as the accompanying 10M word ``dev'' dataset, that we use as a test split.
While the validation split is used to perform the hyperparameter optimization,\footnote{This choice is dictated by the following logic. A hyperparameter sweep can be viewed as a form of optimization. Therefore we would consider using the \texttt{dev} split from BabyLM-2 as a violation of the rules of the challenge. Of course, it means that we trained only on 95\% of the tokens, and could potentially improve our results further.}
the test split is used solely for the purpose of reporting the final cross-entropy loss.
Each dataset is composed of six files, corresponding each to a different type of (English) language that a child is likely to be exposed to, such as transcribed child-directed speech, children’s books, subtitles, or simple Wikipedia.
The relative fractions of these files differ slightly between, on the one hand, the train and validation splits and, on the other, the test split, which is therefore slightly out of distribution.

We have experiment with FineWeb-Edu dataset, \citep{lozhkov2024fineweb-edu}, but have observed that models trained on BabyLM-2 dataset reach better BLiMP scores.

\paragraph{Training.} 
The teacher models are pretrained using the \texttt{Trainer} class from the HuggingFace Transformers library, using the hyperparameters listed in \cref{tab:distillation_hyperparameters}. For the distillation, we use the modified trainer from the original BabyLlama \citep{baby-llama-repo}, with either two or three teachers.
We use the AdamW optimizer \citep{AdamW}, with a cosine schedule for the learning rate and 600 warm-up steps.
The pretraining hyperparameters have been optimized using a coarse-grained scan, with each parameter being varied independently. The distillation hyperparameters $\alpha$ and $T$ were optimized similarly, while holding the pretraining parameters fixed.

All models share the same Byte-Pair Encoding (BPE) tokenizer with a vocabulary size of 16000 trained on the training split of BabyLM-2 dataset.

\begin{table}[h]
\centering
\begin{tabular}{ll}
\toprule
\textbf{Hyperparameter} & \textbf{Value} \\
\midrule
Learning rate & \num{7e-4} \\
Number of epochs & 8 \\
Batch size & 128 \\
Weight decay & 5 \\
\midrule
Distillation $T$ & 1 \\
Distillation $\alpha$ & 0.5 \\
\bottomrule
\end{tabular}
\caption{Training and distillation hyperparameters of BabyLlama-2.}
\label{tab:distillation_hyperparameters}
\end{table}

\paragraph{Hyperparameter Sweep.}
To exclude the possibility that the student model BabyLlama-2 outperforms its teachers due to a suboptimal choice of hyperparameters, we have performed a comprehensive sweep for the teachers’ hyperparameters using the W\&B API \citep{wandb}.
We vary the following hyperparameters: the learning rate and its schedule, the Adam parameters ($\beta_1$, $\beta_2$, $\epsilon$), the batch size, the number of epochs and warm-up steps, the weight decay, the maximum gradient norm, and the attention drop-out.
We use the Bayesian Optimization and Hyperband (BOHB) \citep{BOHB} parallel sweep algorithm, which stops badly-performing runs early, and we minimize the validation loss at the last epoch.
Suitable priors are used for each parameters, usually log-normal or log-uniformly distributed around the values obtained from the coarse-grained scan, with the exception of the attention dropout (uniform) and schedule (discrete).
In total, we trained 265 models as part of the sweep, amounting to 26 GPU-days.
While the sweep produced some runs that perform noticeable better than the teachers trained with the parameters in \cref{tab:distillation_hyperparameters}, re-training them from a different initial state, but otherwise with the exact same parameters, lead to models that significantly under-performed compared to the initial teachers.
Due to this lack of stability with respect to the initialization, we decided to use the original teachers for the distillation procedure.

\paragraph{Benchmarks.}
We evaluate the performance of the teacher and student models on the benchmarks suggested by the organizers of the BabyLM challenge.
Those include zero-shot benchmarks --- such as BLiMP \citep{warstadt-etal-2020-blimp-benchmark}, which focuses on linguistic knowledge in English, and EWoK 
\citep{ivanova2024elementsworldknowledgeewok}, focusing on world knowledge --- as well as the suite of fine-tuning benchmarks SuperGLUE \citep{wang2020superglue} about language understanding.
For the latter, the fine-tuning hyperparameters are optimized using a separate sweep for each task (totalling 1293 runs and 37 GPU-days). The optimal parameters, listed in \cref{tab:fine-tuning-hyperparameters}, differ significantly from the suggested defaults. See \cref{app:superglue} for further discussion.

\paragraph{Baseline models.}
The organizers of {BabyLM-2} have provided two baseline models:
LTG-BERT~\citep{ltg-bert}, (encoder-only) and BabyLlama \citep{timiryasov2023baby} (decoder).
Both models were re-trained by the challenge organizers on both the 10M and 100M word datasets.
LTG-BERT modifies the original BERT architecture by utilizing pre-norm varaint of transformer, GEGLU feed-forward layer and disentangling positional information from token embeddings.
The highest performing solution of the 2023 edition of the BabyLM challenge, ELC-BERT~\cite{elc-bert} is based on this architecture.
On the other hand, BabyLlama (the highest-performing decoder model) uses the standard LLaMA architecture \citep{llama}, but a modified training procedure, following a similar approach to the one presented here. However, in contrast to BabyLlama-2, it was distilled from two larger teachers with two different architectures (GPT and Llama), and had six times less parameters.
Since BabyLlama-2 aims to demonstrate the validity of the ensemble distillation method itself, it uses same-size, homogeneous models in order to remove potential confounding factors.
When evaluating BabyLlama on the SuperGLUE benchmarks, we fine-tune it again using the hyperparameters reported in~\citep{timiryasov2023baby}, and successfully reproduce all of its scores.

\section{Results}
\begin{table*}[htbp!]
    \small
    \centering
    \begin{tabular}{l|ccccc}
        \toprule
        \textbf{Model} & \textbf{BLiMP (filtered)} & \textbf{BLiMP (supplement)} & \textbf{EWoK} & \textbf{SuperGLUE} & \textbf{Macro-average} \\
        \midrule
        BabyLlama-2 (run 1) & $\mathbf{73.2}$ & $63.1$ & $50.6$ & $69.3$ & $64.0$ \\
        \quad Teacher 1 & $71.9$ & $61.8$ & $50.6$ & $61.2$ & $61.3$ \\
        \quad Teacher 2 & $72.1$ & $62.9$ & $50.1$ & $69.5$ & $63.6$ \\
        BabyLlama-2 (run 2) & $71.8$ & $\mathbf{63.4}$ & $\mathbf{51.5}$ & $\mathbf{70.2}$ & $\mathbf{64.2}$ \\
        \quad Teacher 1 & $70.9$ & $62.9$ & $50.4$ & $67.6$ & $62.9$ \\
        \quad Teacher 2 & $70.5$ & $62.4$ & $51.1$ & $68.4$ & $63.1$ \\
        Sweep’s best ckpt.\ & $72.2$ & $60.7$ & $50.1$ & $68.4$ & $62.9$ \\
        \midrule
        BabyLlama (10M) & $69.8$ & $59.5$ & $50.7$ & $63.3$ & $60.8$ \\
        LTG-BERT (10M) & $60.6$ & $60.8$ & $48.9$ & $60.3$ & $57.7$ \\
        \midrule
        BabyLlama (100M) & $73.1$ & $60.6$ & $\mathbf{52.1}$ & $69.0$ & $63.7$ \\
        LTG-BERT (100M) & $69.2$ & $\mathbf{66.5}$ & $51.9$ & $68.4$ & $64.0$ \\
        \bottomrule
    \end{tabular}
    \caption{Summary of the model scores (in $\%$) across the considered benchmarks. The best scores overall and within the \texttt{strict-small} track (10M words maximum) are highlighted.}
    \label{tab:scores-summary}
\end{table*}

\begin{figure*}[htbp!]
\centering
\includegraphics[width=0.95\textwidth]{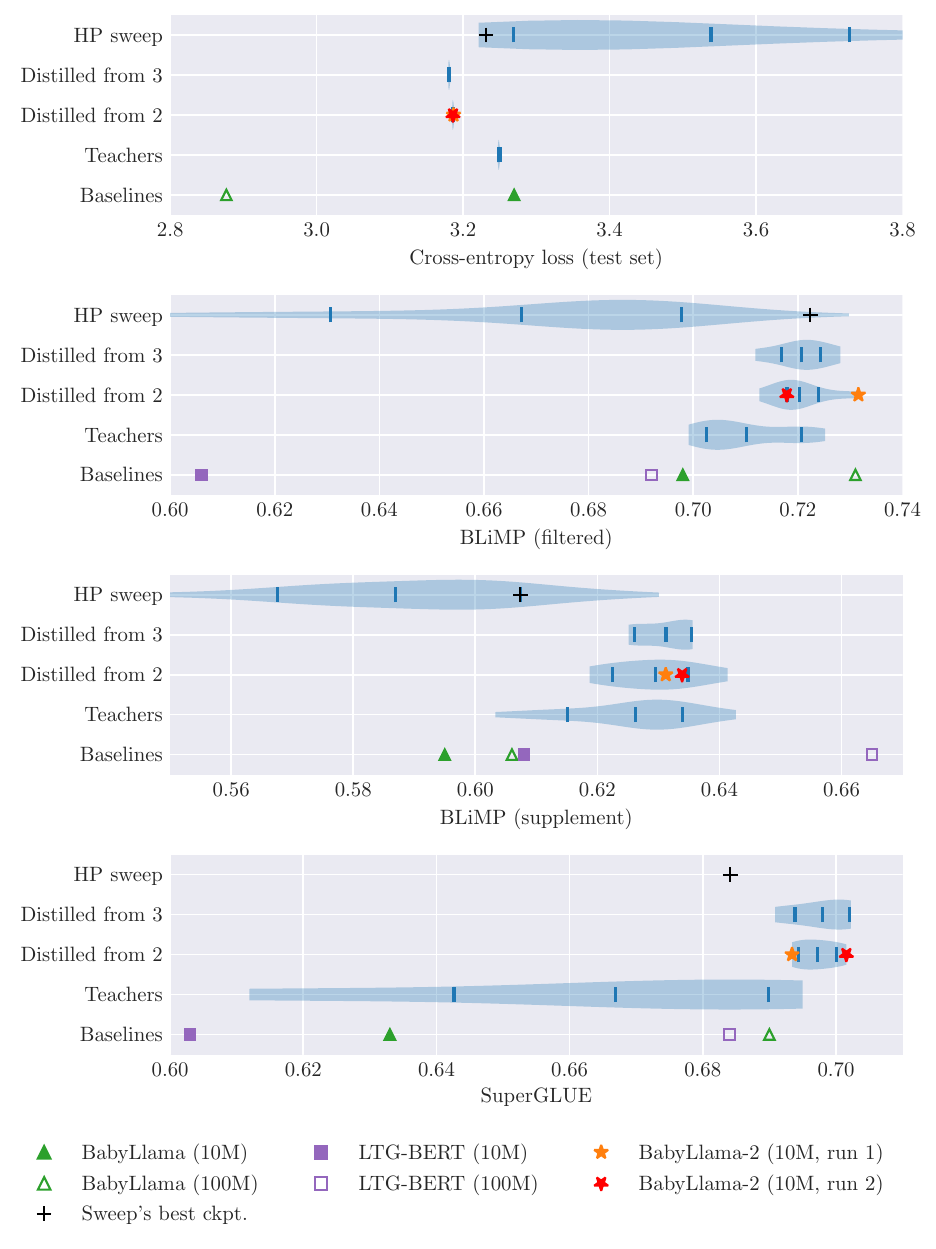}
\caption{Comparison of the models for each evaluation metric, in the form of violin plots, with ticks denoting the mean and $\pm 1$ standard deviation. The baselines are denoted by square and triangle markers, the submitted model (BabyLlama-2) by stars, and the best checkpoint from the entire hyperparameter sweep by a cross. BabyLlama (100M) and LTG-BERT(100M) were trained on the 100M dataset.}
\label{fig:scores}
\end{figure*}

\begin{figure}[h]
\centering
\includegraphics[width=\columnwidth]{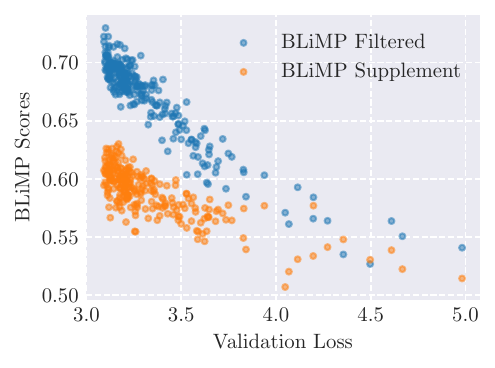}
\caption{BLiMP scores (averaged over all sub-tasks) as a function of the validation loss. Every circle represents a model from the hyperparameter sweep.}
\label{fig:blimp_vs_loss}
\end{figure}

\Cref{fig:scores} summarizes the performance of the models considered in \cref{sec:experimental-setup} with respect to the various evaluation metrics: the cross-entropy loss evaluated on the held-out test set, the BLiMP scores for the ``filtered'' and ``supplement'' subsets of evaluation tasks, and the mean SuperGLUE score.
The EWoK benchmark is not shown, since the performance of our models and of the baselines trained on 10M words is consistent with random chance, hinting that all these models have extremely limited world knowledge, if any.
\paragraph{Distributions.}
Violin plots are used in order to quantify the variability across model initializations, with a minimum of 4~runs per model.
Each subplot shows a different metric, with the $y$-axis listing the various models considered: the teacher models, pretrained without distillation; the student models pretrained with either two or three teachers; the baseline models for the 2024 BabyLM challenge; and the 265 models from the hyperparameter sweep.
No violin is shown for baseline models, since they do not have an associated distribution. Similarly, running fine-tuning benchmarks for all the models from the sweep would have been computationally prohibitive, therefore the SuperGLUE distribution associated with the sweep is not present, with only the best checkpoint being shown.
\paragraph{Models of interest.}
Instead of, or in addition to the distributions, the performance of various models of interest is plotted using markers.
This includes the baseline models, denoted by triangles for BabyLlama and squares for LTG-BERT, with filled markers for baselines pretrained on the 10M word dataset and empty markers for the 100M one.
We also indicate with stars the two BabyLlama-2 models that have been submitted to the 2024 edition of the BabyLM challenge.
Finally, the cross denotes the best model from the entire sweep, as quantified by its validation loss.
The detailed numerical results for the models of interest are listed in \cref{tab:scores-summary}, and \cref{tab:superglue} further details the SuperGLUE scores of the two submitted BabyLlama-2 checkpoints.
\paragraph{Cross-entropy.}
The cross-entropy loss is by far the cleanest metric, with a standard deviation across initializations much smaller than the difference between models.\footnote{The much larger standard deviation for the sweep comes from including all runs (including early and badly performing runs) instead of just the best runs. The relevant quantity for the sweep is therefore the edge of the distribution. The ``best'' model is not always located on this edge, since the validation loss does not correlate perfectly with the test loss or the benchmark scores.} It shows a clear and gradual improvement between the teacher models, the student models trained from two teachers, and those trained from three teachers, although we note that there are diminishing returns when going from two to three teachers. Even with only two teachers, the improvement is larger that what can be achieved through the hyperparameter sweep. However, looking at the BabyLlama baseline\footnote{The cross-entropy loss is not shown for the LTG-BERT baseline, since it is an encoder-only model trained using masked language modelling, and as such its loss is not comparable to the one discussed here.}, it is clear that this improvement is nowhere near the one resulting from using a ten-fold larger dataset.
\paragraph{Benchmarks.}
The scores on the two BLiMP task sets show a similar trend, but with a significantly higher variability across runs. Nonetheless, we can see that the distilled models not only do better than the non-distilled ones, but they tend to achieve this performance more reliably. This is to be contrasted with the performance regression observed after re-training the best model from the sweep.
Another interesting observation is that despite its much lower cross-entropy loss, the BabyLlama baseline pretrained on 100M words only performs on par with the best BabyLlama-2 model trained on 10M words on the ``filtered'' subset of tasks, and significantly underperforms on the ``supplement'' subset.
The results are sensibly similar for the SuperGLUE fine-tuning benchmarks, although with much larger variance among the teacher models. Here, again, the distilled models perform more consistently, and they even beat the two baseline models pretrained on the 100M word dataset.
\paragraph{Relation between loss and benchmark performance.}
The models trained during the hyperparameter sweep allow us to access the relation between the validation loss and BLiMP scores.
First, we observe that the loss on our 0.5M validation set correlates with the loss on the held-out test set with $R^2=0.999$.
Second, as can be seen from \cref{fig:blimp_vs_loss}, the validation loss explains a significant portion of the variance of the scores: $R^2=0.86$ for for BLiMP Filtered and $R^2=0.6$ for BLiMP Supplement.

\paragraph{Discussion.}
The results presented in \cref{fig:scores} demonstrate that ensemble distillation from homogeneous teacher models leads to enhanced and more consistent performance across various benchmarks. Notably, BabyLlama-2 often matches or surpasses models pretrained on datasets that are ten times larger. This indicates that the distillation process effectively leverages the knowledge from multiple teachers to compensate for limited data.
In addition, the performance of distilled models is consistently as good as, or better than the one of non-distilled models, even when optimizing the hyperparameters of the latter.
Therefore, the effect observed in \citet{timiryasov2023baby} cannot be solely the result of badly-tuned hyperparameters, and persists even when the student and teachers share the same size and architecture.
However, this effect can be difficult to see on the benchmark scores, which are much noisier than the cross-entropy loss. This variability is made particularly evident when looking at the different ordering of the two submitted BabyLlama-2 models across different benchmarks.
\paragraph{Limitations}
The scalability of the ensemble distillation approach to larger datasets and more substantial model sizes remains unexplored. It is unclear whether the observed benefits will persist or diminish as the scale of data and model parameters increases. 
Additionally, the exact origin of the improvements from distillation-pretraining remains unclear.

\section{Conclusions}
In this study, we prioritized investigating the robustness of the distillation approach over architectural modifications or dataset curation. Our findings demonstrate that a 345M parameter model, distillation-pretrained on 9.5M words, outperforms models of the same size and architecture pretrained in the usual way. We carried out a systematic analysis to exclude the possibility that the performance gains were due to a single fortunate initialization or suboptimal teacher model hyperparameters. Through an extensive hyperparameter sweep and the training of multiple teacher and student models, we established that distillation-pretraining consistently yields superior performance.

Our results indicate that distillation-pretraining is an effective method for achieving high performance without the need for meticulous hyperparameter tuning, at least within the data-limited regime. The scalability of this approach to larger datasets and model sizes, as well as its applicability to other modalities, remains an open research question.

\section*{Acknowledgements}
We thank Thea Aarrestad, Erik Dam, Troels C.\ Petersen, Oleg Ruchayskiy and Raghavendra Selvan for valuable discussions.
This work was supported by a research grant (VIL57416) from VILLUM FONDEN.
The work of IT was partially supported by the Carlsberg foundation, and by
the European Union's Horizon 2020 research and innovation program under the Marie Sklodowska-Curie grant agreement No. 847523 `INTERACTIONS'.
Computational resources for this work were partially provided by the SCIENCE AI Centre of Copenhagen University.

\appendix

\section{SuperGLUE Fine-tuning}
\label{app:superglue}
The SuperGLUE suite of benchmarks consists of a number of fine-tuning tasks related to language understanding. Since they involve further model training, the scores crucially depend on the chosen fine-tuning hyperparameters. In \cref{tab:fine-tuning-hyperparameters}, we list the hyperparameters used to fine-tune all our models on the SuperGLUE tasks. These parameters were identified using the BabyLlama-2 checkpoint by performing a separate sweep for each task, and then re-starting the fine-tuning with rounded parameters, in order to check the stability of the found parameters. We have observed that they work well with other model checkpoints, including different versions of BabyLlama-2 and teacher models, suggesting that our hyperparameter selection is robust across different model initializations and pretraining objectives (but not model sizes, since the original BabyLlama had different optimal hyperparameters) and is not overfitted to a specific model or task. The detailed SuperGLUE scores of the two BabyLlama-2 checkpoints submitted to the 2024 BabyLM challenge are reported in \cref{tab:superglue}.
\begin{table*}[htb!]
    \small
    \centering
    \begin{tabular}{l|cccccc}
    \toprule
    Task & Max.\ learning rate & Batch size & Num.\ epochs & Weight decay & Schedule & Warm-up steps \\
    \midrule
    CoLA & \num{1e-5} & 32 & 10 & 0.15 & linear & 600 \\
    SST-2 & $2\cdot 10^{-6}$ & 24 & 2 & 5 & constant & 200 \\
    MRPC & $1\cdot 10^{-5}$ & 1 & 2 & 2 & cosine & 500 \\
    QQP & $4.5\cdot 10^{-6}$ & 32 & 6 & 2 & linear & 500 \\
    MNLI(-mm) & $1\cdot 10^{-5}$ & 32 & 2 & 1 & linear & 500 \\
    QNLI & $5\cdot 10^{-6}$ & 32 & 2 & 0.3 & cosine & 200 \\
    RTE & $1\cdot 10^{-5}$ & 2 & 2 & 10 & cosine & 200 \\
    BoolQ & \num{2e-5} & 8 & 1 & 0.1 & cosine & 200 \\
    MultiRC & $1\cdot 10^{-5}$ & 8 & 2 & 2 & cosine & 500 \\
    WSC & $2\cdot 10^{-6}$ & 1 & 24 & 0.4 & cosine & 500 \\
    \bottomrule
    \end{tabular}
    \caption{List of the hyperparameters selected when fine-tuning BabyLlama-2 on the various SuperGLUE tasks. We do not use early-stopping, since it interfered with BOHB’s own early-stopping mechanism. The random seed is 12 for all runs.}
    \label{tab:fine-tuning-hyperparameters}
\end{table*}

\begin{table}[htbp!]
    \small
    \centering
    \begin{tabular}{l|cc}
    \toprule
    Task & Run 1 & Run 2 \\
    \midrule
    CoLA (MCC) & 34.9 & 31.4 \\
    SST-2 & 85.8 & 83.5 \\
    MRPC ($F_1$) & 82.2 & 83.8 \\
    QQP ($F_1$) & 84.1 & 84.3 \\
    MNLI & 74.4 & 74.3 \\
    MNLI-mm & 75.3 & 76.4 \\
    QNLI & 83.3 & 83.2 \\
    RTE & 54.7 & 61.2 \\
    BoolQ & 65.9 & 63.4 \\
    MultiRC & 64.4 & 64.9 \\
    WSC & 57.7 & 65.4 \\
    \bottomrule
    \end{tabular}
    \caption{Detailed scores (in $\%$) of the two BabyLlama-2 models on the SuperGLUE tasks. Unless specified otherwise, the listed score is the accuracy. Hyperparameters were optimized for run~1, and then transferred to run~2.}
    \label{tab:superglue}
\end{table}

\section{Scaling Model Size}
 \label{sec:family}
We performed initial experiments using a small, 16M version of the model, with the same vocabulary size of 16,000; hidden size 256; intermediate size 1024; 8 layers and 8 attention heads.
This model can be fully trained in a few minutes but already achieves decent benchmark scores.

To understand the relationship between model size and data requirements, we conducted additional experiments with our 16M and 345M models. We trained these models on random subsets of the 100M word dataset, ranging from 1M to 100M words each (without re-tuning the hyperparameters).
\begin{figure}[htbp!]
\centering
\includegraphics[width=\columnwidth]{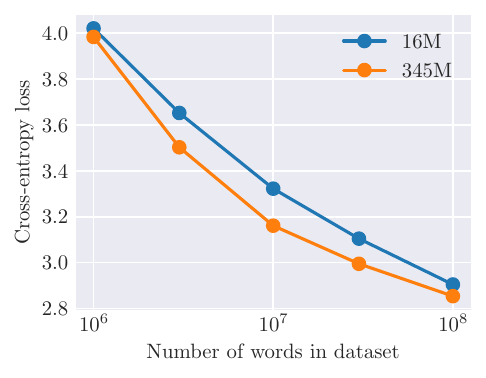}
\caption{Cross-entropy loss as a function of dataset size for 16M and 345M models.}
\label{fig:loss_vs_dataset}
\end{figure}
Figure \ref{fig:loss_vs_dataset} illustrates how the loss decreases as the dataset size increases for both the 16M and 345M models. The 345M model consistently outperforms the 16M model across all dataset sizes, demonstrating that larger models can more efficiently utilize data, hence justifying our choice of the 345M architecture for the final BabyLlama-2 model.

\bibliography{custom}

\end{document}